\documentclass{article}
\usepackage{spconf,amsmath,graphicx,hyperref,indentfirst,multirow,booktabs,makecell,enumitem,caption,subcaption,adjustbox}
\usepackage{hyperref}
\usepackage{etoolbox}
\usepackage{etoolbox}
\apptocmd{\thebibliography}{\setlength{\itemsep}{0pt}}{}{}

\title{Learning Question-Aware Keyframe Selection with Synthetic Supervision for Video Question Answering}
\name{Author(s) Name(s)\thanks{Thanks to XYZ agency for funding.}}
\name{Minchan Kwon$^\dagger$ \qquad Hyounguk Shon$^\dagger$ \qquad Junmo Kim\thanks{$^{\dagger}$ Equal contribution.}}
\address{Korea Advanced Institute of Science and Technology, South Korea}

\begin{document}

\maketitle
\begin{abstract}
Large multimodal models (LMMs) have recently demonstrated remarkable performance in video question answering (VideoQA), yet reasoning over video remains challenging due to high inference cost and diluted information. Keyframe selection offers efficiency and sharper reasoning but suffers from sparse supervision and redundant frame choices when relying only on image–text similarity. We present a question-aware keyframe selection framework with two components: pseudo keyframe labels derived from LMMs that provide informative supervision and a coverage regularization that promotes diverse, complementary evidence across time. Experiments on NExT-QA show that our method significantly improves accuracy, especially for temporal and causal question types, establishing keyframe selection as an effective and learnable module for VideoQA.

\end{abstract}

\section{Introduction}
\label{sec:intro}

% 최근 LLM에 시각정보 이해를 추가한 LMM이 VideoQA에서 뛰어난 성능을 보이고 있음. (e.g., InstructBLIP, LLaVA) 그러나 맥락 시퀀스가 길어질 경우 추론 비용이 크게 증가할 뿐만 아니라, 긴 맥락 속에서 QA에 유관한 비디오 프레임을 찾아내는 능력이 제한적인 것으로 밝혀지고 있음. 특히 이러한 단점은 scene-/episode-level VideoQA에서 두드러짐. 따라서 우리는 키프레임 선택 접근방식을 통해 샘플링 시 question과 관계가 높은 프레임을 출하여 효율성 향상 및 중요 프레임에 집중. 

% VideoQA를 위한 keyframe selection 모듈 학습의 난점은 keyframe 선택에 사용할 수 있는 supervision signal이 매우 희소하다는 것임. 먼저 키프레임에는 ground truth 라는 개넘이 없다. 따라서 기존 연구에서는 사전학습된 image-text model (CLIP)에서 추출한 prompt-frame similarity score 를 기반으로 선택하는 방법이 일반적이었다. 그러나 이 방식은 프레임간의 관계가 고려되지 않으므로 causal, temporal 같은 선후관계의 이해가 필요한 문제에서 키프레임 선택의 정확도가 낮은 문제가 있다. Causal, temporal 질문에서는 ITM이 capture하지 못하는 video-question relationship이 존재하기 때문이다. 
% 또 하나의 패러독스는 이러한 문제를 해결하기 위해서는 keyframe 선택 모듈이 주어진 QA 문제에 대한 고수준의 이해를 할 수 있어야 하는 문제점이 있다. 
% 뿐만 아니라 similarity score 기반의 접근 방법은 highest score 기준으로 선택할 경우 인접한 프레임들이 선택되어 keyframe에 담긴 정보가 redundant 해지는 문제점이 있다. 

% 이에 따라 우리는 similarity score에 coverage regularization을 적용하는 방법과, LMM을 이용해 synthetic keyframe label을 생성하는 방법을 제안한다. 

Large multimodal models (LMMs) that extend language models with visual understanding have recently achieved strong performance on Video Question Answering (VideoQA). Systems such as InstructBLIP and LLaVA further narrow the gap between open-ended language queries and long, visually rich content~\cite{instructblip,llava,video-llama}. Despite these gains, reasoning on sequences of images in VideoQA remains challenging. As the number of frames grows, inference cost scales substantially, and the model’s ability to localize question-relevant visual evidence within lengthy streams degrades~\cite{framevoyager}. These limitations are particularly pronounced in scene-/episode-level settings, where answers depend on sparse, widely separated moments that are easily drowned out by redundant context.

An effective solution lies in \emph{keyframe selection}, reducing a video to a compact, question-aware subset of frames that captures evidence while eliminating irrelevant content. Such selection promises a dual benefit. First, it improves \emph{efficiency} by shortening the tokenized visual sequence the LMM must process. Second, it sharpens \emph{reasoning} by concentrating attention on informative moments rather than forcing the model to sift through long stretches of distractors.

However, learning a keyframe selector for VideoQA faces a fundamental obstacle: supervision is extremely sparse. There is no universally agreed ground-truth set of keyframes, and typical VideoQA datasets provide only question--answer pairs without temporal annotations. Furthermore, the discrete nature of keyframe selection (i.e., choosing the top-k frames) is inherently non-differentiable, making end-to-end training challenging. To circumvent this, previous works have resorted to noisy gradient approximation techniques~\cite{gcg,vidf4}, while some utilize reinforcement learning to directly optimize for selection policies~\cite{framevoyager}. However, these methods often introduce significant algorithmic complexity and training instability. Consequently, many resort to framewise \emph{image--text} similarity (e.g., CLIP~\cite{clip}) to rank frames against the question~\cite{keyvideollm,gcg,vidf4}, despite its own set of critical shortcomings: 
\begin{enumerate}
\item \textbf{Temporal/causal relations are ignored.} Framewise matching treats frames independently and fails to capture inter-frame dependencies central to temporal and causal reasoning; datasets such as NExT-QA, AGQA, and CLEVRER explicitly probe these abilities~\cite{nextqa,grunde2021agqa,cleverer}. Questions such as ``\emph{Why} did the character pick up the phone?'' or ``What happens \emph{after} the door opens?'' require relations across frames that static image--text alignment cannot capture.
\item \textbf{Redundancy under top-score selection.} Selecting the top-$k$ frames by similarity often yields temporally adjacent near-duplicates, leading to poor coverage of the diverse evidence a question may require (e.g., setup, action, consequence, before/after states, multiple participants). Diversity-/coverage-oriented selection objectives (e.g., AKS~\cite{tang2025adaptive}) directly address this issue.
\item \textbf{A supervision paradox.} To pick the \emph{right} frames, a selector needs a high-level understanding of the video--question interaction that we hope to enable downstream---yet such understanding is not directly supervised by standard VideoQA labels~\cite{nextqa}.
\end{enumerate}
\begin{figure*}[!t]
    \centering
    \includegraphics[width=\linewidth]{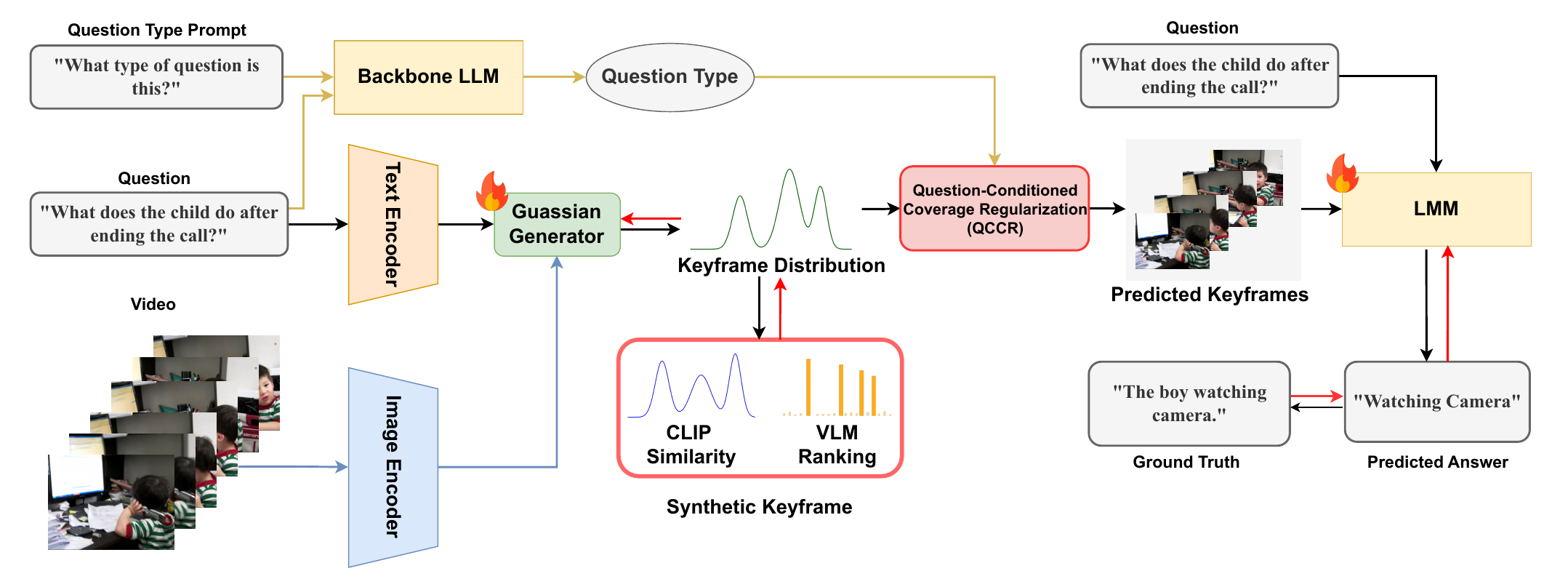}
    \caption{An overview of our proposed question-aware keyframe selection framework. The model processes the video with an image encoder and the question with a text encoder. The question embedding is then used in two parallel streams: (1) it is fed into a Gaussian Generator to create an initial keyframe distribution, and (2) it is used to generate synthetic keyframes via CLIP similarity and VLM ranking, which serve as weak supervision. To adapt the selection process, the LMM's own backbone is re-purposed with a prompt to determine the question type. This information guides the Question-Conditioned Coverage Regularization (QCCR) module, which refines the initial distribution. Finally, the predicted keyframes are passed to the downstream LMM to generate the final answer.}
    \label{fig:figure1}
\end{figure*}
\textbf{Our approach.}
We address these challenges with two complementary ideas:
\begin{itemize}
\item \textbf{Synthetic keyframe label generation.} To overcome the scarcity of supervision, we prompt a strong LMM to produce coarse rationales (e.g., timestamps or frame indices) for its answers and transform these signals into \emph{pseudo keyframe labels}. Although noisy, these labels encode higher-level video--question relations (including temporal and causal cues) that pure image--text matching misses. We further stabilize learning by aggregating multiple prompts/views and aligning selection quality with downstream answer correctness.
\item \textbf{Question-conditioned coverage-regularized selection.} Starting from a similarity-based scoring backbone (which can leverage CLIP or the vision encoder of an LMM), we introduce a \emph{question-conditioned} coverage regularization term that explicitly discourages redundant choices and promotes diversity across semantics and time. Concretely, the selector is rewarded for picking frames that collectively ``cover'' more video frames, rather than saturating a single moment with near-duplicates. 
\end{itemize}

Together, these components yield a question-conditioned selector that (i) reduces inference cost by passing far fewer frames to the LMM and (ii) preserves or improves answer quality by focusing on evidence-bearing moments, with particular gains on temporal/causal categories. 

Our contributions are summarized as follows: 
% \begin{enumerate}
% \item We formulate \emph{question-aware keyframe selection module} for video question answering, targeting both efficiency and reasoning quality.
% \item We introduce a \emph{question-conditioned coverage regularization} selection that mitigates redundancy and encourages complementary evidence across time depending on semantics.
% \item We propose a practical pipeline to obtain \emph{synthetic keyframe supervision} from an LMM, injecting high-level video--question relations into the selector without manual temporal annotations.
% \item We demonstrate our method using NExT-QA benchmark~\cite{nextqa}, with particular gains on causal and temporal question types, and provide ablations isolating the effects of coverage regularization and synthetic supervision.
% \end{enumerate}

% By coupling principled diversity with informative pseudo-labels, our approach turns keyframe selection from a brittle heuristic into a learnable, question-aligned module that helps LMMs reason over a long frame sequence at a fraction of the cost.

\begin{enumerate}
\item We design a \emph{question-aware keyframe selection module} for video question answering, aiming to improve both efficiency and reasoning quality.
\item We develop a method to derive \emph{synthetic keyframe supervision} from an LMM, injecting high-level video--question relations into the selector without requiring manual temporal annotations.
\item We introduce a \emph{question-conditioned coverage regularization} that reduces redundancy and promotes complementary evidence across time, conditioned on question semantics.
\item We evaluate our method on the NExT-QA benchmark~\cite{nextqa}, showing notable gains on causal and temporal questions, and conduct ablations isolating the effects of coverage regularization and synthetic supervision.
\end{enumerate}

By coupling principled diversity with informative pseudo-labels, our approach transforms keyframe selection from a heuristic process into a learnable, question-aligned module, enabling LMMs to reason over long frame sequences at a fraction of the computational cost.
\section{METHOD}
\label{sec:method}

\subsection{Framework Overview}
Our goal is to learn a keyframe selector that identifies $N$ keyframes from a video 
$V = \{x_0, x_1, \ldots, x_T\}$ and a question $Q$.  
We adopt the Gaussian Generator framework from GCG~\cite{gcg} to design the keyframe selector.  
The Gaussian Generator takes input video embeddings from the CLIP image encoder and question embeddings from the text encoder.  
It outputs $K$ Gaussian masks $g = \{g_1, \ldots, g_K\}$.  
The keyframe probability distribution is defined as:
\[
p = \text{Norm}\left(\Sigma_{k=1}^{K} g_k\right) \qquad p \in [0,1]\,,
\]
where $\text{Norm}(\cdot)$ normalizes the value to $[0,1]$. The Gaussian masks are trained with pseudo timestamp labels $w = [w_1, \ldots, w_K]$ using MSE loss: 
\[
L_{mse} = \frac{1}{K}\sum_{k=1}^K (g_k - w_k)^2.
\]
We select the top-$N$ keyframes from the Gaussian masks.  
These frames are passed through the Q-former~\cite{instructblip} and then into the LMM.  
The final answer is trained with a VQA loss:
\[
L_{vqa} = \mathcal{L}_{\text{VQA}}(A, \hat{A}),
\]
where $A$ is the ground truth answer and $\hat{A}$ is the model output.  

\subsection{Synthetic Keyframe Supervision (SKS)}
We propose a synthetic keyframe generation framework to provide temporal grounding.  
Following GCG, we first compute CLIP similarities and select the top-$N$ frames as supervision.  
However, CLIP similarity assigns similar scores to visually similar scenes, regardless of information content.  
It also ignores temporal order, making it weak for causal reasoning.  
To address this, we leverage LMMs for keyframe supervision.  
LMMs capture temporal and causal relations through attention.  
We design prompts that guide the LMM to select frames relevant to answering the question.  
The system prompt given to the LMM is designed around four prioritized selection rules. 
\begin{enumerate}
    \item Informative first - Choose frames that explicitly show the entities/actions/clues referenced in the question or answer choices. 
    \item Non-redundancy - If multiple frames are nearly identical, keep the earliest, clearer one. 
    \item Temporal logic - When sequence matters (before/after cues), preserve correct order. 
    \item Fallback to uniform - Only when none of the frames satisfy (1)-(3), select four frames that are as evenly spaced as possible across the video. 
\end{enumerate}
The LMM is prompted to generate the indices of the selected frames. Finally, the model is trained with both CLIP-based supervision and synthetic keyframe supervision. 

\subsection{Question-conditioned Coverage Regularization}
Different question types require different keyframes.  
We propose Question-conditioned Coverage Regularization (QCCR) to adapt keyframe selection at test time.  
Coverage regularization~\cite{tang2025adaptive} is used to enhance the post-hoc selection of keyframes given the scores associated with each video frame. At test time, the coverage regularization implicitly maximizes 
\[
L_{cov} = \sum_{t\in\mathcal{I}}{s(Q,x_t)} + \lambda(Q) \cdot c(\mathcal{I}) \,,
\]
where $s(Q,x_t)$ is the score of the $t$-th frame, $\mathcal{I}$ is the set of keyframe indices, and $c(\mathcal{I})$ is the coverage regularization. 
The regularization strength $\lambda$ controls how uniform the keyframe indices should be distributed across time. 

If the regularization is either too weak or too strong, the selected keyframes fails to capture the information required for answering the question.
Therefore, coverage regularization needs to be applied adaptively depending on the semantics of the question. 
To this end, we propose to condition $\lambda(Q)$ to the question and adapt to the type of the question. 
This requires that the keyframe selector must have the capability to understand the question prompt. 
Therefore, we re-use the LLM backbone with a query prompt containing the question to predict its type.  
For each question type, we assign a predefined coverage regularization strength.  
\section{EXPERIMENT}
\label{sec:experiment}
% Please add the following required packages to your document preamble:
% \usepackage{booktabs}
\begin{table}[]
\centering
\begin{tabular}{@{}l|cccc@{}}
\toprule
         & \multicolumn{4}{c}{NExT-QA}               \\
         & Descriptive & Temporal & Causal & Average \\ \midrule
Co-Mem   & 54.4        & 50.0     & 45.9   & 48.5    \\
HCRN     & 54.0        & 49.3     & 47.1   & 48.9    \\
HGA      & 57.8        & 49.1     & 48.1   & 50.0    \\
IGV      & 59.6        & 51.7     & 48.6   & 51.3    \\
HQGA     & 59.4        & 52.3     & 49.0   & 51.8    \\
B2A      & 58.3        & 49.0     & 47.4   & 49.6    \\
VCSR     & 62.3        & 51.5     & 53.0   & 54.1    \\
VGT      & 67.3        & 54.5     & 52.8   & 55.7    \\
Raformer & 67.8        & 57.7     & 58.2   & 59.6    \\
TranSTR  & 70.0        & 60.2     & 59.7   & 61.5    \\
GCG      & 69.0        & \textbf{74.4}     & 67.8   & 70.4    \\
SeViLA   & \textbf{80.8}        & 66.4     & \textbf{71.9}   & \underline{71.5}    \\ \midrule
Ours     & \underline{77.8}        & \underline{71.5}     & \underline{71.6}   & \textbf{73.3}    \\ \bottomrule
\end{tabular}

\caption{Performance comparison with state-of-the-art methods on the NEXT-QA benchmark. All methods are evaluated on accuracy across three question categories. Our method achieves the highest average accuracy, demonstrating its overall effectiveness.}
\label{tab:main_table}
\end{table}

\begin{figure}[!t]
    \centering
    \includegraphics[width=\columnwidth]{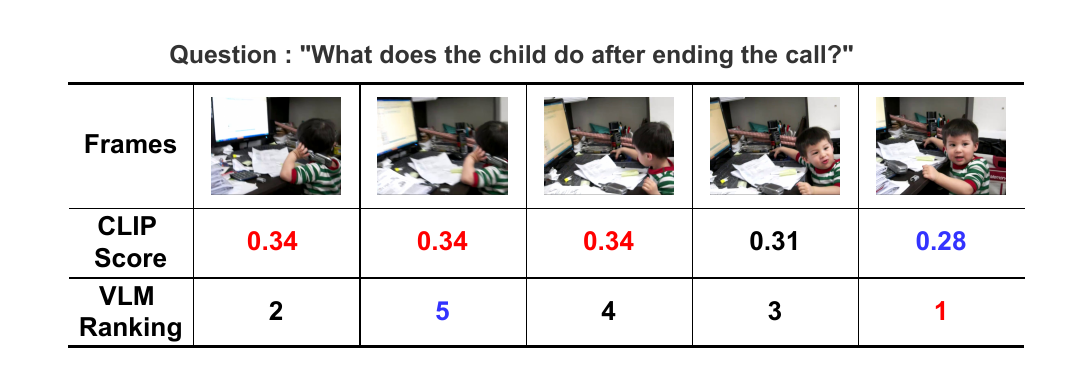}
    \caption{A qualitative example illustrating the effectiveness of VLM Ranking over CLIP similarity. For the question ``What does the child do after ending the call?", frames with high CLIP scores are often redundant while the VLM Ranking correctly identifies the temporally relevant frames that occur after the call has ended.}
    \label{fig:figure2}
\end{figure}
\subsection{Experiment Setting}
\textbf{Dataset.}
We use the NExT-QA dataset~\cite{nextqa}.  
NExT-QA contains 5.4k videos with an average length of 44s and 52k QA pairs.  
The dataset covers three types of questions: \emph{descriptive}, \emph{temporal}, and \emph{causal}.  

\textbf{Baselines.}
% We compare our method with a range of representative VideoQA models.  
% HCRN~\cite{hcrn} leverages hierarchical conditional reasoning across clips and frames.  
% HGA~\cite{hga} models question–video interactions with a heterogeneous graph.  
% IGV~\cite{igv} introduces invariant grounding for robust video–language alignment.  
% HQGA~\cite{hqga} enhances video reasoning via hierarchical question–guided attention.  
% B2A~\cite{b2a} improves causal reasoning by bridging background and answer spaces.  
% TranSTR~\cite{transtr} performs adaptive temporal rationalization for frame selection.  
% SeViLA~\cite{sevila} uses LMMs for both pseudo-label generation and answering, with extra pre-training on TSG datasets.  
% We compare our method with a range of representative VideoQA models, including HCRN~\cite{hcrn}, HGA~\cite{hga}, IGV~\cite{igv}, HQGA~\cite{hqga}, B2A~\cite{b2a}, TranSTR~\cite{transtr}, and SeViLA~\cite{sevila}. These baselines cover various approaches from hierarchical reasoning to adaptive frame selection.
We compare our method against representative VideoQA baselines. One group of methods focuses on relational reasoning, using hierarchical structures~\cite{hcrn} or heterogeneous graphs~\cite{hga,hqga}. Others improve performance through robust video-language alignment~\cite{igv}, causal reasoning~\cite{b2a}, or adaptive frame selection~\cite{transtr}. We also compare against SeViLA~\cite{sevila}, a strong LMM-based method.
Different from previous works, our model employs CLIP~\cite{clip} for weak supervision and a lightweight Gaussian Generator to learn multiple masks.  

\textbf{Experiment Details.}
We set $T=32$ frames per video and select $N=4$ keyframes.  
For pseudo-label generation, we use Qwen2.5-VL~\cite{Qwen2.5-VL} as the LMM, which is a state-of-the-art open-source LMM. 
All hyperparameters for GCG training are kept consistent with the original settings.  
We adopt InstructBLIP~\cite{instructblip} as the LMM backbone and EVA-CLIP~\cite{evaclip} as the text–image encoder for the Gaussian Generator.  

\subsection{Main Results}
Table~\ref{tab:main_table} reports the main results.  
Our method achieves the highest average score, demonstrating a robust and well-balanced performance across all question types. 
This demonstrates that our method performs robustly across question types, rather than achieving high scores due to being particularly strong on specific questions.
Compared to our direct baseline, GCG, there was a significant performance improvement in both descriptive and causal questions.
This highlights that our proposed synthetic keyframes and QCCR are effective at capturing complex semantic and causal relationships that simple similarity-based selection misses. 
Furthermore, while the state-of-the-art method SeViLA excels in the descriptive category, our model not only outperforms it on temporal questions but also secures a better overall average. Considering that SeViLA employs an LMM with billions of parameters during keyframe selection, our method, which utilises only a Gaussian generator in the form of a linear layer, also offers computational advantages.

\subsection{Qualitative Results}
Figure~\ref{fig:figure2} displays the frames for specific questions from the NExT-QA dataset, alongside their corresponding CLIP scores and VLM Rankings.
The CLIP score assigns higher scores to frames containing “child” and “call” elements mentioned in the question, whereas the VLM Ranking assigns the highest ranking to frames occurring after the call has ended, reflecting the causal and temporal nature of the question.
Conversely, it assigned the lowest ranking to the unhelpful phone-answering scene.
This demonstrates that relying on CLIP for ground truth keyframes presents difficulties in causal scenarios.
Our method, which utilizes both VLM and CLIP grounding results, can blend the strengths of both approaches.

\begin{table}[t]
\centering
\resizebox{0.9\linewidth}{!}{
\begin{tabular}{@{}c|cccc@{}}
\toprule
                                & \multicolumn{4}{c}{NExT-QA}               \\
                                & Descriptive & Temporal & Causal & Average \\
\midrule
Ours                            & \textbf{77.8}        & \textbf{71.5}     & \textbf{71.6}   & \textbf{73.3}    \\
\makecell{w/o QCCR}       
                                & 75.9        & 67.9     & 70.9   & 71.5    \\
\makecell{w/o SKS} 
                                & 69.0        & 74.4     & 67.8   & 70.4    \\ 
\bottomrule
\end{tabular}
}
\caption{Ablation study of our core components on the NEXT-QA dataset. The performance drop in both cases confirms that each component makes a crucial contribution to the final performance.}
\label{tab:ablation}
\end{table}
\subsection{Ablation Study}
To disentangle the contributions of each component, we conducted ablation experiments on the NExT-QA dataset (Table~\ref{tab:ablation}).
Removing QCCR caused a clear drop, especially on temporal and causal questions. Without it, the model over-selected redundant frames, reducing evidence diversity.
Eliminating synthetic keyframe supervision led to the sharpest decline, with descriptive accuracy falling from 77.8 to 69.0. Pure similarity-based selection failed to capture semantics, focusing on visually dominant but uninformative frames.
Interestingly, the \emph{w/o SKS} scores higher on temporal questions than our full method. We hypothesize this is because the simple similarity-based selection aligns well with certain patterns in the temporal questions, but its poor performance elsewhere highlights a lack of generalizability.

\section{CONCLUSION}
\label{sec:conclusions}
In this work, we addressed the key challenges of supervision scarcity and frame redundancy in keyframe selection for VideoQA.
We introduced two complementary solutions: Synthetic Keyframe Supervision (SKS) to provide high-level semantic guidance, and Question-conditioned Coverage Regularization (QCCR) to ensure temporal diversity. 
% This enhances the utility of keyframe selection by resolving the issues of insufficient ground truth data and narrow coverage. 
We showed that with proper supervision and regularization, keyframe selection can be elevated from a simple retriever to a stronger learnable component, achieving state-of-the-art performance in NExT-QA. 
% Future work could explore extending to long videos or investigating more sophisticated methods for generating synthetic keyframe supervision.

\bibliographystyle{IEEEbib}
\bibliography{refs}

\end{document}